\documentclass[10pt,twocolumn,letterpaper]{article}

\PassOptionsToPackage{numbers, compress}{natbib}

\usepackage[pagenumbers]{arxiv} 

\usepackage{url}
\usepackage{dsfont}
\usepackage{afterpage}

\usepackage{arydshln}
\usepackage{multirow} 
\usepackage{colortbl} 
\usepackage{bbm}
\usepackage{algorithm}
\usepackage{algpseudocode}                  
\usepackage{setspace}                       
\usepackage{lipsum} 
\usepackage{cuted}
\usepackage{capt-of}
\usepackage{wrapfig}
\usepackage{pifont} 

\usepackage{mathtools} 
\usepackage{bm} 
\usepackage{soul} 
\usepackage{nicefrac}
\usepackage{csquotes}

\usepackage{graphicx}
\usepackage{booktabs}
\usepackage{float}
\usepackage[table]{xcolor}

\usepackage{duckuments}

\definecolor{ourslight}{RGB}{235,242,250}
\definecolor{oursdark}{RGB}{222,234,247}
\makeatletter
\newcommand*{\myfnsymbol}[1]{%
  \ensuremath{%
    \ifcase#1\or
      \text{\ding{41}}\or
      1\or
      2\or
      3\or
      4\or
      5\or
      6\or
      7\else
      8\fi
  }%
}
\def\@fnsymbol#1{\myfnsymbol{#1}}
\def\thempfootnote{\myfnsymbol{\c@mpfootnote}}
\makeatother

\definecolor{cornellred}{rgb}{0.7, 0.11, 0.11}
\definecolor{cadmiumgreen}{rgb}{0.0, 0.42, 0.24}
\definecolor{aliceblue}{rgb}{0.91, 0.94, 0.97}
\definecolor{darkblue}{rgb}{0.83, 0.89, 0.97}
\definecolor{Red7}{rgb}{0.941, 0.243, 0.243}
\definecolor{Green7}{RGB}{55, 178, 77}
\definecolor{Blue9}{rgb}{0.21098,0.493,0.74}

\sethlcolor{aliceblue}

\definecolor{cvprblue}{rgb}{0.21,0.49,0.74}
\usepackage[pagebackref,breaklinks,colorlinks,allcolors=cvprblue]{hyperref}

\title{
From RGB Generation to Dense Field Readout: \\
Pixel-Space Dense Prediction with Text-to-Image Models

}

\author{ 
\vspace{-25pt}\\
\textbf{Zanyi Wang}$^{1}$\hspace{3mm}
\textbf{Xin Lin}$^{1}$\hspace{3mm} 
\textbf{Haodong Li}$^{1}$\hspace{2mm}  \\
\textbf{Dengyang Jiang}$^{2}$  \hspace{2mm}
\textbf{Yijiang Li}$^{1}$\hspace{2mm} \\
$^1$UCSD~ $^2$HKUST \\[1mm]
 \hspace{-4mm} \url{https://github.com/xmz111/ReChannel}
 }

\makeatletter
\g@addto@macro\@maketitle{
 \vspace{-1.0em}
\begin{figure}[H]
\setlength{\linewidth}{\textwidth}
\setlength{\hsize}{\textwidth}
\centering
\includegraphics[width=\textwidth]{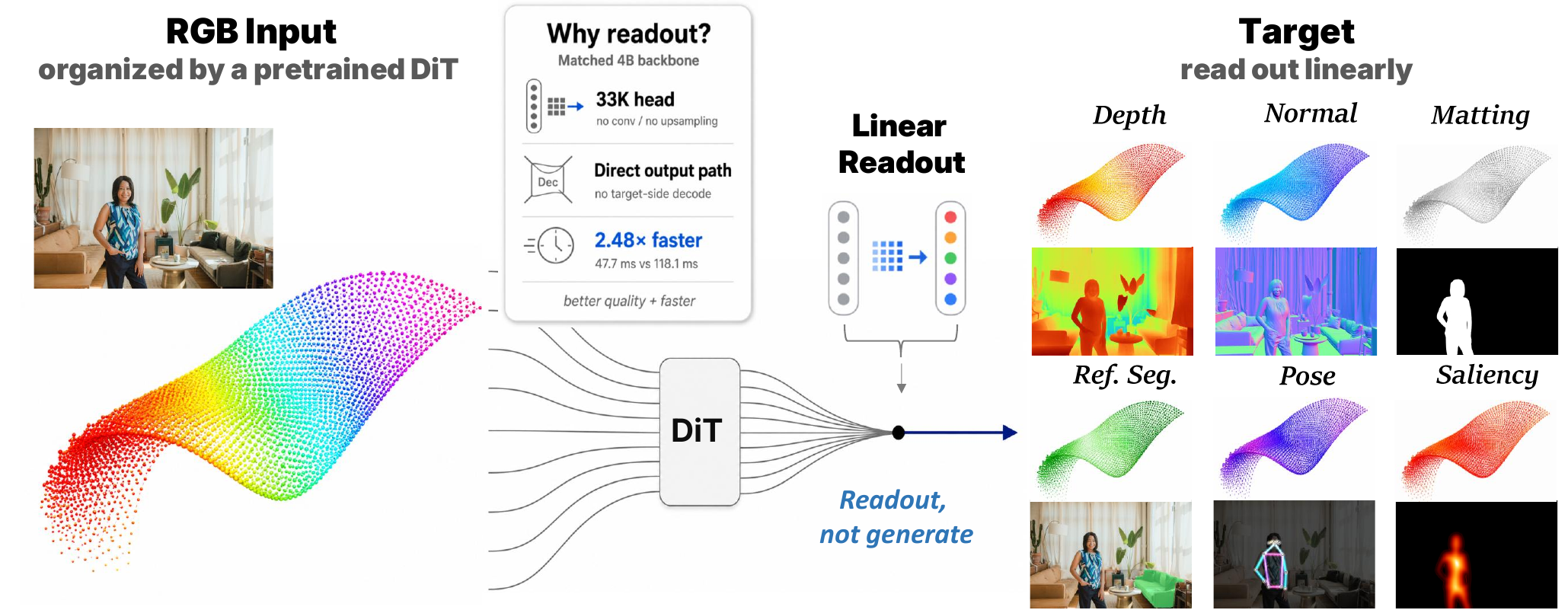} 
\caption{
    \textbf{ReChannel: readout, not generation.}
    A pretrained DiT organizes RGB inputs into a patch-aligned spatial token field, so
    dense prediction becomes reading out task-native quantities on the same image plane
    rather than reconstructing an RGB-style target. Each output patch is a spatial carrier
    in the DiT lattice; the readout reinterprets its channels from RGB appearance to
    task-native fields---depth, surface normals, matting, referring segmentation, pose, and
    saliency. Since these targets are evaluated as pixel-space fields, target-side VAE
    decoding is unnecessary.
    }

\label{fig:paradigm}

\end{figure}
}
\makeatother

\begin{document}

\maketitle

\begin{abstract}
Large-scale text-to-image models are attractive backbones for dense prediction because RGB generation pretraining learns rich semantic, structural, and geometric priors. Existing generative and editing approaches reuse these priors by casting dense prediction as target generation: annotations such as depth, normals, alpha mattes, masks, and heatmaps are encoded into an RGB-trained VAE latent space and decoded back as image-like targets. We argue this inherits more of the generative output interface than dense prediction requires: unlike RGB synthesis, dense prediction asks for pixel-correct, task-native fields on the same image plane, not new RGB content to be rendered.
Our key observation is that a pretrained DiT already organizes RGB inputs
through a patch$\to$token$\to$patch lattice on the image plane, so each token
indexes a fixed output patch whose channels can carry task-native quantities
instead of RGB appearance. We instantiate this as \textbf{ReChannel}: we keep the VAE encoder for the DiT's input distribution but drop the
target-side decoder, adapt the frozen DiT with task LoRA, and map each token to
its $p\times p\times K_t$ pixel-space patch through a shared token-local linear
head---about 33K parameters, no spatial mixing.
Using FLUX-Klein, we evaluate on six dense prediction tasks and over a dozen benchmarks. This minimal interface sets new state-of-the-art on trimap-free matting, KITTI depth, and referring segmentation, and stays competitive on normals, saliency, and pose. In a matched 4B setting it is more accurate and 2.48$\times$ faster than an edit-plus-latent-decode counterpart---dense perception can benefit from generative pretraining without inheriting its output interface.
\end{abstract}

\begin{figure}[t]
    \centering
    \includegraphics[width=\columnwidth]{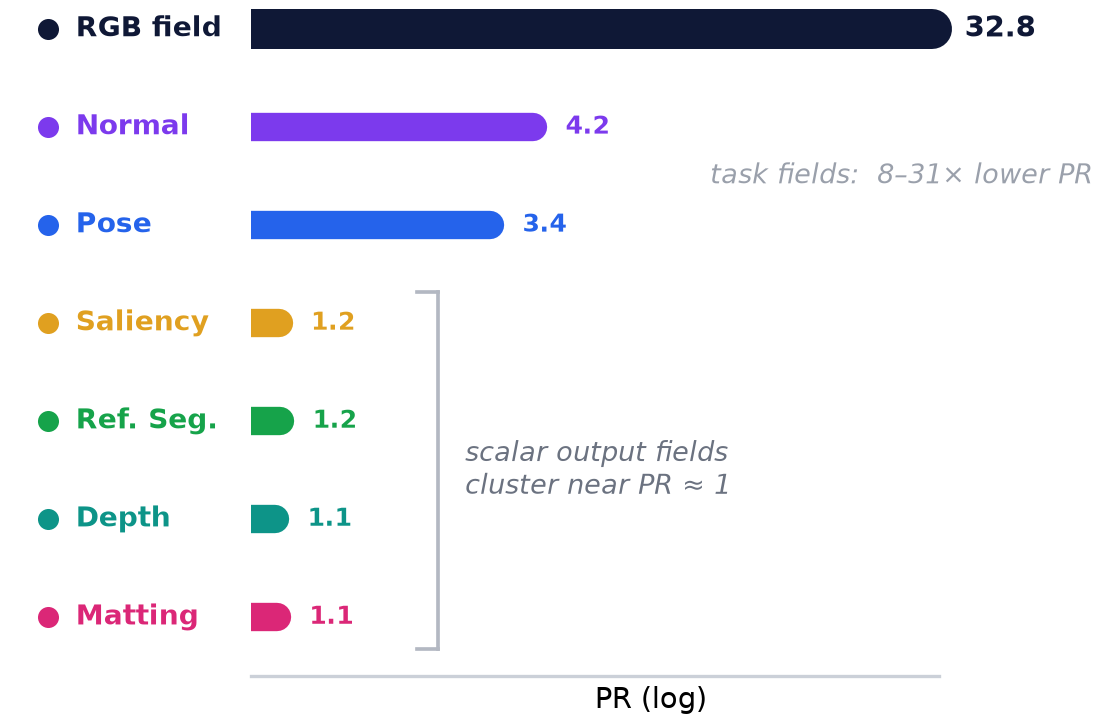}
    \caption{
    Participation ratio (PR) of the token field: the RGB input field (32.8) is high-dimensional, while task-adapted fields collapse to compact subspaces (1.1--4.2), motivating a token-local linear readout rather than a generative decoder. Per-task PR values are illustrative; the linear head's sufficiency is shown empirically in Sec.~\ref{sec:exp_ablation}.
    }
    \label{fig:compact_fields}
\end{figure}

\section{Introduction}
\label{sec:intro}

\vspace{-0.2cm}

RGB-based dense prediction begins with a simple constraint: the RGB image is
the only visual observation from which pixel-aligned fields such as geometry,
masks, mattes, and heatmaps are estimated. The dominant discriminative
paradigm addresses this problem by pairing pretrained visual encoders with
task-specific decoders, increasingly benefiting from foundation
representations such as DINO, DINOv2, and SAM
~\cite{caron2021emerging,oquab2023dinov2,kirillov2023segment}. These models
provide strong RGB-conditioned visual features, but the dense target is still
obtained through learned task decoders that translate the representation into
pixel-space fields~\cite{ranftl2021vision,cheng2022masked,yang2024depth,xu2022vitpose}.

Text-to-image models offer a different route at larger generative scale~\cite{flux2klein2026,z-image,qwenimage,seedream4}. Their
RGB synthesis objective learns rich image-formation priors over appearance,
semantics, structure, and layout, which recent generative dense predictors
have reused across depth, surface normals, segmentation, and matting
~\cite{ke2024repurposing,fu2024geowizard,wang2025deforming,he2025lotus}.
These methods have rapidly evolved from iterative diffusion to deterministic
one-step prediction and image-to-image editing
~\cite{xu2025matters,he2025lotus,wang2025editor,shi2025edit2perceive}. Yet most VAE-based generative or editing approaches keep the same target-side
output interface: the dense target is represented as an image-like or latent
target, passed through the latent space of an RGB-trained VAE, and decoded back
to the final pixel-space prediction. This is natural for RGB synthesis, but
indirect for dense prediction: task-native targets must be fit into an
RGB-trained latent space, while the supervision, prediction, and evaluation of
interest are pixel-space fields. As illustrated in Fig.~\ref{fig:paradigm}, we
ask whether dense prediction should inherit this target-side RGB reconstruction
path, or instead use the generative model only as an RGB image-formation prior
from which task-native fields can be read out directly.

We argue that the right interface should reflect the asymmetry between RGB
generation and dense prediction. RGB generation must reconstruct a complex
appearance field of color, texture, illumination, and high-frequency detail,
whereas dense prediction re-expresses this evidence as task-native pixel-space
fields such as geometry, opacity, or heatmaps, not RGB appearance. This is visible in our diagnostic (Fig.~\ref{fig:compact_fields}): the RGB input token field is high-dimensional, while task-adapted fields occupy much more compact subspaces. This does not mean dense tasks are visually simple; it means the pretrained RGB field can be adapted into compact, task-aligned readout subspaces. We verify this directly in Sec.~\ref{sec:exp_ablation}: a target-side VAE decoder is an inherited rendering module, not a necessary dense prediction interface.

Our key observation follows from the patch structure of modern image
Transformers. ViTs establish a spatial lattice of image-patch tokens, and DiTs
extend it into a patch-to-token-to-patch transformation for
generation~\cite{dosovitskiy2020image,peebles2023scalable}. An output patch is
thus a spatial carrier: in generation its channels are interpreted as RGB
appearance, but for dense prediction the same carrier can hold task-native
quantities. Dense prediction still requires task-specific semantics, but its final outputs
are task-native fields on the image plane, not images to be rendered through a
generative decoder.

\begin{figure*}[t]
\centering
\includegraphics[width=\textwidth]{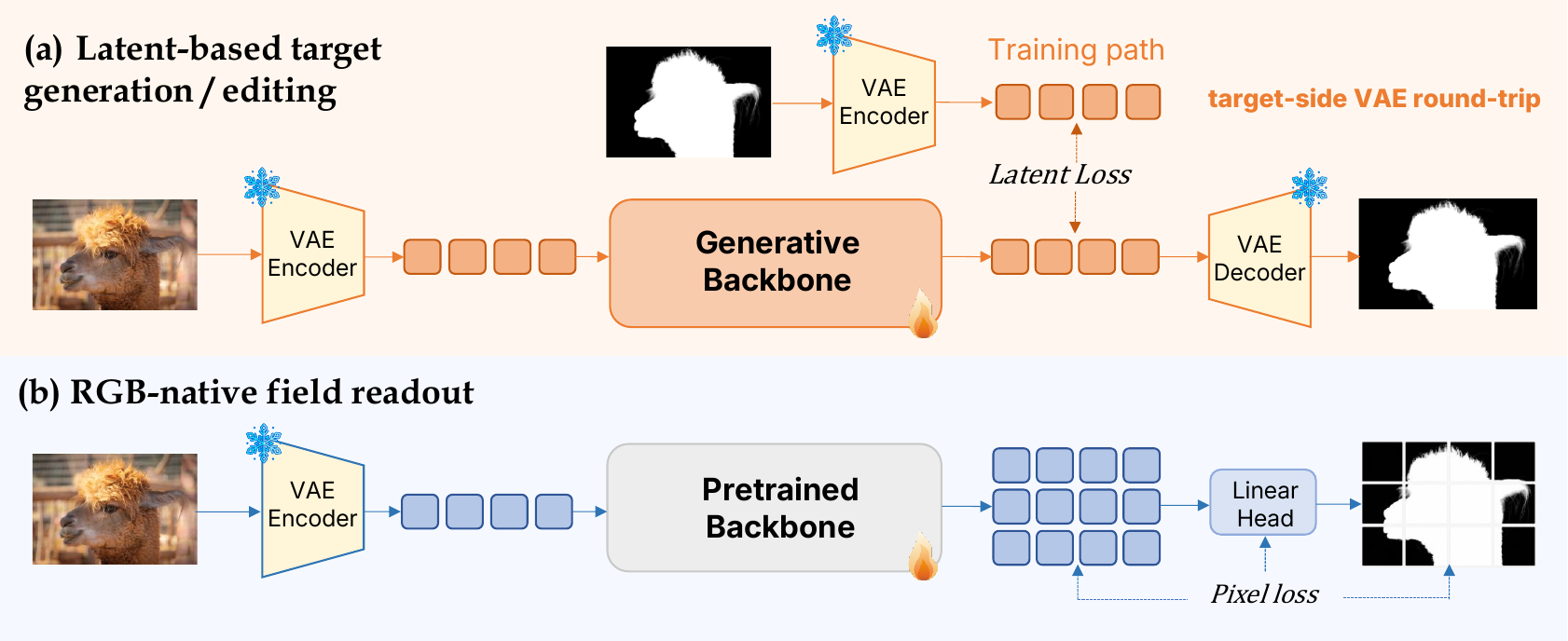}
\vspace{-2mm}
\caption{
    \textbf{ReChannel vs.\ target-side generation.}
    Existing generative and editing interfaces treat dense targets as image-like outputs:
    the target is encoded into an RGB-trained VAE latent, supervised there, and decoded back
    through a target-side reconstruction path. ReChannel instead keeps the RGB input interface
    but drops the target-side path: the pretrained backbone acts as an RGB-native field
    organizer, and a lightweight linear readout maps its spatial tokens directly to
    task-native pixel-space patches under pixel supervision---field readout rather than target
    generation.
    }
\label{fig:main}
\vspace{-2mm}
\end{figure*}

As shown in Fig.~\ref{fig:main}(b), we instantiate this insight as
\textbf{ReChannel}. The pretrained VAE encoder is retained to
preserve the latent input distribution expected by the DiT, while dense targets
never enter the VAE. Lightweight task LoRA adapts the generation-pretrained
token field toward the target semantics, and a shared token-local linear
projection maps each adapted token to its corresponding
$p\times p\times K_t$ target patch. Since the DiT already organizes the RGB
observation as a patch-aligned spatial field, the readout need only reinterpret
the channels of each spatial carrier from RGB appearance to task-native
quantities. The adapted token field therefore provides the spatial structure,
allowing direct field readout without a high-capacity spatial decoder or
target-side reconstruction path.

We validate this recipe on six dense prediction tasks and more than a dozen
benchmarks, where a single token-local readout reaches state-of-the-art results
across tasks as different as continuous geometry, high-frequency matting,
language-conditioned segmentation, and multi-channel pose heatmaps. Controlled
ablations under a matched backbone show these gains come from the interface, not
from head capacity or scale: a $13\times$ larger head does not help, training
from scratch collapses, and latent, VAE-decoded, or edit outputs are all less
accurate and up to $2.48\times$ slower. The right use of generative pretraining,
then, is not to generate the target but to read out the dense field it already
organizes.

\noindent\textbf{Our contributions are threefold:}
\begin{itemize}
\item \textbf{Rechanneling, not generating.} We introduce \textbf{ReChannel}: an output patch is a spatial carrier whose channels can be reinterpreted as task-native fields rather than RGB, reframing dense prediction as rechanneling, not generation.
\item \textbf{A minimal interface.} A single recipe---frozen T2I backbone, lightweight adaptation, and a token-local linear readout with no spatial decoder---handles geometry, masks, language-conditioned segmentation, and heatmaps in one shared form.
\item \textbf{Strong, efficient results.} Across six dense prediction tasks and a dozen-plus benchmarks, ReChannel reaches the state-of-the-art frontier while running up to 2.48$\times$ faster than an edit-plus-latent-decode counterpart.
\end{itemize}

\section{Related Work}

\textbf{From task decoders to unified visual backbones.}
Dense prediction has traditionally been feature-to-field decoding: depth,
surface normals, matting, segmentation, and pose each define their own target
representation, supervision, and output head. With the rise of visual foundation models, this landscape has shifted toward stronger and more transferable visual representations. Self-supervised encoders such as DINO and DINOv2 provide dense visual features, SAM provides promptable segmentation priors, and dense prediction systems such as DPT, Mask2Former, Depth Anything, and ViTPose show how scalable visual backbones are commonly paired with task-specific or task-unified decoders for depth, segmentation, and pose~\cite{caron2021emerging,oquab2023dinov2,kirillov2023segment,ranftl2021vision,cheng2022masked,yang2024depth,xu2022vitpose}. These methods show the strength of discriminative pretraining, but largely retain a decoder-centric view: features become dense outputs through learned task heads, upsampling, or spatial decoders.

\textbf{Generative priors for dense perception.}
Text-to-image generative models have recently been repurposed as dense prediction backbones. Marigold and GeoWizard adapt latent diffusion priors to monocular depth and surface normals, showing that RGB synthesis pretraining captures useful geometric structure~\cite{ke2024repurposing,fu2024geowizard}. GenPercept broadens this view to general dense perception and suggests that deterministic one-step adaptation is often better aligned with perception than iterative denoising~\cite{xu2025matters}. Lotus further moves toward direct annotation prediction, while recent editor-based methods formulate dense perception as image-to-image generation or editing~\cite{he2025lotus,wang2025editor,shi2025edit2perceive,shan2026cyclegen, zhao2026geostream}. These works establish generative pretraining as a powerful source of dense visual priors. However, representative VAE-based approaches still treat dense targets as objects to be generated: annotations are encoded into an RGB-trained latent space and decoded back through a VAE. Our work separates the generative prior from this target-side interface: we reuse the prior but do not decode the target.

\textbf{Output interfaces and patch-indexed generation.}
A recent generalist direction pushes the generative view further: image-generation pretraining is argued to serve as a universal interface, with diverse vision tasks expressed as RGB image outputs~\cite{gabeur2026image}. Orthogonally, the ViT/DiT line treats images as patch-indexed token grids, and pixel-space generators such as JiT show that such grids can model image-plane quantities without a VAE tokenizer or decoder~\cite{dosovitskiy2020image,peebles2023scalable,li2026back}. This VAE-free interface, however, has so far been shown only for generation; deterministic dense prediction still routes its targets through a VAE. This raises a narrower question: if a T2I backbone provides a strong RGB-native prior, must dense outputs still be rendered through an image-generation interface?

\section{Method}
\label{sec:method}

\subsection{Dense Prediction as Field Readout}
\label{sec:method_readout_view}

We study the output interface between a pretrained text-to-image backbone and
deterministic dense prediction. Existing generative approaches expose dense
targets through an image-generation interface: the target is produced as an
image or latent and decoded back to pixels (Fig.~\ref{fig:main}a). This is natural for RGB synthesis,
but dense prediction is supervised and evaluated as task-native quantities on
the image plane---so we ask what the output interface actually needs to be.

We instead treat the spatial token grid of the T2I transformer as the readout substrate for dense prediction. After lightweight adaptation, each token is mapped to its local pixel-space target patch by a shared linear projection—no generative decoder, no high-capacity spatial head. The token field, not the head, carries the spatial structure; the head only reads it out.

This gives \textbf{ReChannel} (Fig.~\ref{fig:main}b): the RGB
input pathway is preserved, but dense targets never enter the VAE. Each adapted
token is read out into its pixel-space target patch, so the output interface is
task-native rather than generative.

\subsection{Adapting the RGB Field}
\label{sec:method_lora}

Given an input image $x$, we follow the original input pathway of the pretrained T2I model and encode the RGB image with its VAE encoder. This keeps the input in the distribution and coordinate system on which the generative backbone was pretrained. We then run the DiT backbone in a deterministic zero-noise mode, with the backbone weights frozen and only task-specific LoRA parameters $\Delta_t$ inserted:
\begin{equation}
Z_t = F_{\theta+\Delta_t}\big(\mathrm{VAE}_{\mathrm{enc}}(x); \sigma=0, c_t\big),
\label{eq:field_adaptation}
\end{equation}
where $F_\theta$ is the pretrained T2I transformer, $c_t$ is the text condition, and $Z_t=\{z^t_{ij}\}$ denotes the resulting spatial token field for task $t$.

This design keeps the RGB side of the generative model intact. The VAE encoder is used only as the pretrained RGB input interface, allowing the DiT to operate on the kind of latent field it was trained to model. The dense target, however, is not required to pass through the corresponding generative output interface. Depth, normals, alpha mattes, masks, and heatmaps are supervised and evaluated as pixel-space fields, so we use LoRA to adapt the RGB field toward the target semantics and leave the final conversion to a direct readout.

In this view, LoRA only reshapes how the pretrained RGB field is organized toward the target semantics; it does not build the dense output. Each task uses its own adapter and readout, and the dense output is left entirely to the readout in Sec.~\ref{sec:method_head}.

\subsection{Token-Local Pixel Readout}
\label{sec:method_head}

After task adaptation, each spatial token is mapped directly to a local dense patch by a shared token-local linear projection:
\begin{equation}
\hat{Y}^{t}_{ij}
=
\mathrm{reshape}\left(W_t z^t_{ij}+b_t\right),
\quad
\hat{Y}^{t}_{ij}\in\mathbb{R}^{p\times p\times K_t}.
\label{eq:token_local_readout}
\end{equation}
Here $p$ is the patch size and $K_t$ the number of output channels for task $t$; tiling all patches over the image plane gives the final prediction $\hat{Y}^t$.

The backbone maps an input RGB patch to a token; the readout maps that token to a target patch at the same image location $\Omega_{ij}$. The token plays the same role for the target as for RGB---a spatial carrier indexed to a fixed region---and only its output channels change, from RGB to the $K_t$ task-native channels.

The readout is a single linear map applied independently at every token: no multi-stage decoder and no inter-token spatial mixing. The reshape acts as a fixed sub-pixel projection to pixel space, not a learned spatial decoder. Any spatial structure in the output therefore comes from the adapted token field, not from the head. We call this readout \textbf{ReChannel}: it reinterprets each token's channels from RGB appearance to task-native fields, leaving the spatial carrier to the token field. That a purely token-local linear map suffices is the central empirical claim of this work, which we validate across heterogeneous tasks in Sec.~\ref{sec:exp_main} and probe in the ablations of Sec.~\ref{sec:exp_ablation}.

\begin{table*}[t]
\centering
\caption{
Main results on geometric dense prediction. 
Depth entries report absRel $\downarrow$ / $\delta_1 \uparrow$.
Normal entries report mean angular error $\downarrow$ / $11.25^\circ$ accuracy $\uparrow$.
}
\label{tab:main_geometry}
\small
\setlength{\tabcolsep}{4.2pt}
\begin{tabular}{lccc|ccc}
\toprule
\multirow{2}{*}{Method}
& \multicolumn{3}{c|}{Monocular depth}
& \multicolumn{3}{c}{Surface normals} \\
\cmidrule(lr){2-4}
\cmidrule(lr){5-7}
& NYU & KITTI & ScanNet
& NYU & ScanNet & iBims \\
\midrule
\multicolumn{7}{l}{\emph{Specialized / discriminative methods}} \\
Omnidata v2 (ICCV'21)~\cite{eftekhar2021omnidata}
& -- & -- & --
& 17.2 / 55.5
& 16.2 / 60.2
& 18.2 / 63.9 \\
DAv2 (NIPS'24)~\cite{yang2024depth}
& 0.045 / 0.979
& 0.074 / 0.946
& -- 
& -- & -- & -- \\
DSINE (CVPR'24)~\cite{bae2024rethinking}
& -- & -- & --
& 16.4 / 59.6
& 16.2 / 61.0
& 17.1 / 67.4 \\
\midrule
\multicolumn{7}{l}{\emph{Generative dense prediction methods}} \\
GeoWizard (ECCV'24)~\cite{fu2024geowizard}
& 0.052 / 0.966
& 0.097 / 0.921
& 0.061 / 0.953
& 17.0 / 56.5
& 15.4 / 61.6
& 19.3 / 63.0 \\
Marigold (CVPR'24)~\cite{ke2024repurposing}
& 0.055 / 0.964
& 0.099 / 0.916
& 0.064 / 0.952
& 16.1 / 60.5
& 14.5 / 66.1 
& 16.3 / 68.5 \\
GenPercept (ICLR'25)~\cite{xu2025matters}
& 0.056 / 0.960
& 0.130 / 0.842
& 0.062 / 0.961
& 18.3 / 56.0
& 17.7 / 58.3
& 18.2 / 64.0 \\
Lotus-D (ICLR'25)~\cite{he2025lotus}
& 0.053 / 0.977
& 0.093 / 0.928
& 0.060 / 0.963
& 16.2 / 59.8
& 14.7 / 64.0
& 17.1 / 66.4 \\
Edit2Perc (CVPR'26)~\cite{shi2025edit2perceive}
& \textbf{0.044} / \textbf{0.976}
& 0.079 / 0.945
& 0.049 / 0.973
& 15.7 / \textbf{61.6}
& 14.1 / 66.3
& 15.1 / 70.9 \\
\midrule
\multicolumn{7}{l}{\emph{RGB-native field readout}} \\
\rowcolor{ourslight}
\textbf{ReChannel-4B (Ours)}
& 0.056 / 0.964
& 0.067 / 0.952
& 0.058 / 0.966
& 15.8 / 60.8
& 14.3 / 64.9
& 15.5 / 70.0 \\
\rowcolor{oursdark}
\textbf{ReChannel-9B (Ours)}
& 0.051 / 0.974
& \textbf{0.063} / \textbf{0.959}
& \textbf{0.047} / \textbf{0.976}
& \textbf{15.6} / 61.2
& \textbf{13.9} / \textbf{66.6}
& \textbf{15.1} / \textbf{71.1} \\
\bottomrule
\end{tabular}
\end{table*}

\subsection{Task Instantiations}
\label{sec:method_tasks}

The same readout form is used across tasks. For each task, we train a task-specific LoRA adapter and readout projection, with the output channel dimension $K_t$ and supervision loss chosen according to the target representation. Referring segmentation additionally uses the referring expression as the text condition. Scalar fields such as depth, alpha, and saliency use $K_t=1$; surface normals use $K_t=3$; pose estimation uses multi-channel heatmaps. Continuous regression tasks are trained with their standard pixel-space losses, while binary mask tasks use mask supervision in pixel space. We keep these choices conventional, since the focus of our study is the output interface rather than task-specific loss design.

\begin{table*}[t]
\centering
\caption{
Trimap-free alpha matting on P3M-500.
We report SAD $\downarrow$, MSE$\times 10^3 \downarrow$, and Conn $\downarrow$.
Blank entries indicate that the corresponding metric is not reported in the source paper.
}
\label{tab:main_matting}
\small
\setlength{\tabcolsep}{4.2pt}
\begin{tabular}{lccc|ccc|ccc}
\toprule
\multirow{3}{*}{Method}
& \multicolumn{3}{c|}{P3M-500-P}
& \multicolumn{3}{c|}{P3M-500-NP} 
& \multicolumn{3}{c}{AIM-500 (Zero-shot)}\\
\cmidrule(lr){2-4}
\cmidrule(lr){5-7}
\cmidrule(lr){8-10}
& SAD & MSE & Conn
& SAD & MSE & Conn
& SAD & MSE & Conn\\
\midrule
\multicolumn{9}{l}{\emph{Trimap-free matting methods}} \\
SHM (ACMMM'18)~\cite{chen2018semantic}
& 21.56 & 10.0 & 17.53
& 20.77 & 9.3  & 17.09 
& - & - & - \\
HATT (CVPR'20)~\cite{qiao2020attention}
& 25.99 & 5.4 & 25.29 
& 30.53 & 7.2 & 27.42 
& - & - & - \\
MODNet (AAAI'21)~\cite{ke2022modnet}
& 13.31 & 3.8 & 10.88 
& 16.70 & 5.1 & 13.81 
& - & - & - \\
P3M-Net (ACM    MM'21)~\cite{li2021privacy}
& 8.73 & 2.6  & 13.88
& 11.23 & 3.5 & 12.51 
& 103.03 & 54.47 & 87.29 \\
ViTAE-S (IJCV'23)~\cite{ma2023rethinking}
& 6.24 & 1.5 & 5.86 
& 7.59 & 1.9 & 6.96 
& 112.52 & 60.20 & 43.18 \\
\midrule
\multicolumn{9}{l}{\emph{Generative dense prediction methods}} \\
GenPercept (ICLR'25)~\cite{xu2025matters}
& 9.75 & 2.5 & 8.84
& 12.77 & 2.7 & 10.46 
& 75.5 & 24.20 & 36.74\\
\midrule
\multicolumn{9}{l}{\emph{RGB-native field readout}} \\
\rowcolor{ourslight}
\textbf{ReChannel-4B (Ours)}
& 5.99 & 1.3 & 5.73
& 6.98 & 1.5 & 6.55 
& 43.83 & 19.20 & 42.19\\
\rowcolor{oursdark}
\textbf{ReChannel-9B (Ours)}
& \textbf{5.69} & \textbf{1.2}  & \textbf{5.42}
& \textbf{6.67} & \textbf{1.4} & \textbf{6.23} 
& \textbf{34.90} & \textbf{14.18} & \textbf{32.01}\\
\bottomrule
\end{tabular}
\end{table*}

\begin{table*}[t]
\centering
\caption{
Referring segmentation on RefCOCO-family benchmarks.
We report cIoU (\%) on the standard 8 splits and the average over splits.
}
\label{tab:main_refseg}
\small
\setlength{\tabcolsep}{3.1pt}
\begin{tabular}{lccccccccc}
\toprule
Method
& \multicolumn{3}{c}{RefCOCO}
& \multicolumn{3}{c}{RefCOCO+}
& \multicolumn{2}{c}{RefCOCOg}
& Avg. \\
\cmidrule(lr){2-4}
\cmidrule(lr){5-7}
\cmidrule(lr){8-9}
& val & testA & testB
& val & testA & testB
& val & test
& \\
\midrule

CRIS (CVPR'22)~\cite{wang2022cris}
& 70.5 & 73.2 & 66.1
& 62.3 & 68.1 & 53.7
& 59.9 & 60.4
& 64.3 \\

PolyFormer (CVPR'23)~\cite{liu2023polyformer}
& 74.8 & 76.6 & 71.1
& 67.6 & 72.9 & 59.3
& 67.8 & 69.1
& 69.9 \\

LISA-7B (CVPR'24)~\cite{lai2024lisa}
& 74.9 & 79.1 & 72.3
& 65.1 & 70.8 & 58.1
& 67.9 & 70.6
& 69.9 \\

GLaMM-7B (CVPR'24)~\cite{rasheed2024glamm}
& 79.5 & 83.2 & 76.9
& 72.6 & 78.7 & 64.6
& 74.2 & 74.9
& 75.6 \\

PSALM-1.3B (ECCV'24)~\cite{zhang2024psalm}
& 83.6 & 84.7 & 81.6
& 72.9 & 75.5 & 70.1
& 73.8 & 74.4
& 77.1 \\

OMG-LLaVA-7B (NIPS'24)~\cite{zhang2024omg}
& 78.0 & 80.3 & 74.1
& 69.1 & 73.1 & 63.0
& 72.9 & 72.9
& 72.9 \\

Text4Seg-8B (ICLR'25)~\cite{lan2025text4seg}
& 79.2 & 81.7 & 75.6
& 72.8 & 77.9 & 66.5
& 74.0 & 75.3
& 75.4 \\

READ-7B (CVPR'25)~\cite{qian2025reasoning}
& 78.1 & 80.2 & 73.2
& 68.4 & 73.7 & 60.4
& 70.1 & 71.4
& 71.9 \\

CoPRS-7B (ICLR'26)~\cite{lu2025coprs}
& 81.6 & 85.3 & 79.5
& 75.9 & 80.3 & 69.7
& 76.2 & 76.2
& 78.1 \\

FCLM-7B (CVPR'26)~\cite{yang2026hugging}
& 82.6 & 83.9 & 79.9
& 77.4 & 80.5 & 71.3
& 79.4 & 80.5
& 79.4 \\

\midrule
\rowcolor{ourslight}
\textbf{ReChannel-4B (Ours)}
& 83.8 & 85.6 & 81.6
& 77.9 & 81.6 & 72.5
& 79.2 & 79.8
& 80.3 \\

\rowcolor{oursdark}
\textbf{ReChannel-9B (Ours)}
& \textbf{84.9} & \textbf{86.5} & \textbf{83.7}
& \textbf{79.5} & \textbf{83.0} & \textbf{75.0}
& \textbf{81.2} & \textbf{81.9}
& \textbf{82.0} \\

\bottomrule
\end{tabular}
\end{table*}

\section{Experiments}
\label{sec:exp}

\subsection{Setup}
\label{sec:exp_setup}

We use FLUX-Klein (4B and 9B variants) as the pretrained text-to-image
backbone and freeze it for all main results, training a single LoRA
adapter and a token-local linear head per task with deterministic
$\sigma{=}0$ inference~\cite{flux2klein2026}. All evaluations follow each benchmark's standard
protocol: depth on NYU (Eigen split), KITTI (Garg crop), and ScanNet val
with log-space scale-shift alignment; surface normals on the DSINE
NYU\,/\,ScanNet\,/\,iBims split; trimap-free matting on P3M-500-P,
P3M-500-NP, and zero-shot AIM-500; referring segmentation on the
LISA/GLaMM 8-split RefCOCO/\,+/\,g protocol with cumulative cIoU;
saliency on DUTS-TE and ECSSD with the canonical PySODMetrics
$F_{\max}$; and pose on COCO val with the YOLOv11x detection-box
protocol.

\subsection{Main Results}
\label{sec:exp_main}

\textbf{Geometry: depth and normals (Table~\ref{tab:main_geometry}).}
On monocular depth, ReChannel-9B sets the best absRel on KITTI ($0.063$,
$-0.016$ vs.\ Edit2Perc) and ScanNet ($0.047$, $-0.002$), and trails
Edit2Perc only on NYU ($0.051$ vs.\ $0.044$). On surface normals,
ReChannel-4B is already competitive with the strongest generative baselines
on every split, and ReChannel-9B achieves the best mean angular error on all
three datasets (NYU $15.6^\circ$, ScanNet $13.9^\circ$, iBims
$15.1^\circ$), even though the same token-local linear readout is used
with $K_t{=}3$.

\textbf{Trimap-free matting (Table~\ref{tab:main_matting}).}
Matting is the most boundary-sensitive task in our study and the one
where a target-side VAE round-trip is expected to hurt most. ReChannel-9B
sets a new state of the art on both P3M-500-P (SAD $5.69$) and
P3M-500-NP (SAD $6.67$), improving over the strongest specialized
matting backbone ViTAE-S by $0.55$ and $0.92$ SAD respectively, and
over the strongest generative-prior baseline GenPercept by $4.06$ and
$6.10$. The gap widens out of distribution: on zero-shot AIM-500,
ReChannel-9B reaches $34.90$ SAD, less than half of GenPercept ($75.5$), suggesting
that removing the target-side VAE helps not only in-domain boundaries but also
cross-domain transfer.

\textbf{Referring segmentation (Table~\ref{tab:main_refseg}).}
The same readout, with the referring expression supplied as the
backbone's text condition, handles language-conditioned segmentation
without any LLM branch, mask-proposal head, or task-specific decoder.
ReChannel-4B already outperforms 7B--8B LLM-based baselines (LISA, GLaMM,
Text4Seg) and the strongest recent method FCLM-7B ($80.3$ vs.\ $79.4$
average cIoU). ReChannel-9B reaches $82.0$ average cIoU and is the best on
all 8 RefCOCO splits, indicating that a T2I-pretrained backbone with a
token-local readout already supplies both the spatial grounding and
the segmentation precision that prior work obtains through dedicated
mask heads or LLM decoders.

\textbf{Pose and saliency (Table~\ref{tab:main_pose_saliency}).}
These two tasks test the readout in two extreme regimes: multi-channel
keypoint heatmaps and high-contrast binary masks. On COCO pose,
ReChannel-9B reaches $79.2$ AP, surpassing ViTPose-L by $+0.9$ AP without
any pose-specific architecture; the same $p{\times}p{\times}K_t$
readout is instantiated with a smaller $p$ to match the standard
heatmap resolution. On saliency, ReChannel-9B is the best on both
$F_{\max}$ (DUTS-TE $0.944$, ECSSD $0.968$) and MAE (DUTS-TE $0.018$,
ECSSD $0.017$) across all reported methods.

\begin{table*}[t]
\centering
\caption{
Additional dense prediction tasks.
Pose estimation evaluates structured multi-channel heatmap readout on COCO, while saliency detection evaluates binary mask readout.
Pose entries report AP metrics (\%) on COCO val.
Saliency entries report $F_{\max}\uparrow$ / MAE $\downarrow$.
}
\label{tab:main_pose_saliency}
\small

\begin{minipage}[t]{0.48\textwidth}
\centering
\textbf{(a) COCO human pose estimation}\\[1mm]
\setlength{\tabcolsep}{3.2pt}
\begin{tabular}{lcc}
\toprule
Method & AP & AP$_{50}$ \\
\midrule
HRNet-W48 (CVPR'19)~\cite{sun2019deep}
& 76.3 & 90.8 \\

TokenPose-L (ICCV'21)~\cite{li2021tokenpose}
& 75.8 & 90.3 \\

ViTPose-L (NIPS'22)~\cite{xu2022vitpose}
& 78.3 & 91.4 \\

RTMO-l (CVPR'24)~\cite{lu2024rtmo}
& 74.8 & -- \\

DynPose (CVPR'25)~\cite{xu2025dynpose}
& 75.0 & 90.6 \\

\midrule
\rowcolor{ourslight}
\textbf{ReChannel-4B (Ours)}
& 78.0 & 91.4  \\

\rowcolor{oursdark}
\textbf{ReChannel-9B (Ours)}
& \textbf{79.2} & \textbf{92.7}  \\

\bottomrule
\end{tabular}
\end{minipage}
\hfill
\begin{minipage}[t]{0.49\textwidth}
\centering
\textbf{(b) Saliency detection}\\[1mm]
\setlength{\tabcolsep}{3.0pt}
\begin{tabular}{lccc}
\toprule
Method & DUTS-TE  & ECSSD \\
\midrule
ICON (TPAMI'22)~\cite{zhuge2022salient}
& 0.904 / 0.037  & 0.954 / 0.032 \\

MENet (CVPR'23)~\cite{wang2023pixels}
& 0.917 / 0.028  & 0.957 / 0.031 \\

VST++ (TPAMI'24)~\cite{liu2024vst++}
& 0.887 / 0.033  & 0.949 / 0.029 \\

VSCode-B (CVPR'24)~\cite{luo2024vscode}
& 0.930 / 0.022  & 0.961 / 0.022 \\

EVPv2 (TPAMI'26)~\cite{liu2025explicit}
& 0.921 / 0.026  & 0.967 / 0.022 \\

\midrule
\rowcolor{ourslight}
\textbf{ReChannel-4B (Ours)}
& 0.939 / 0.018  & 0.965 / 0.018 \\

\rowcolor{oursdark}
\textbf{ReChannel-9B (Ours)}
& \textbf{0.944} / \textbf{0.018}  & \textbf{0.968} / \textbf{0.017} \\

\bottomrule
\end{tabular}
\end{minipage}

\end{table*}

\begin{table*}[t]
\centering
\caption{
Diagnostic ablation under a matched 4B backbone (FLUX-Klein 4B, $512^2$, $\sigma{=}0$).
Each italic block isolates one question; every row perturbs a single axis from our configuration.
Normal: mean angular error ($^\circ\downarrow$); Matting: SAD ($\downarrow$) on P3M-500-P / -NP.
Latency is single-stream L40S GPU time.
}
\label{tab:ablation_diagnostic}
\small
\setlength{\tabcolsep}{5.2pt}
\begin{tabular}{lcccccc}
\toprule
\multirow{2}{*}{Configuration}
& \multicolumn{3}{c}{Surface normal ($^\circ\downarrow$)}
& \multicolumn{2}{c}{Matting (SAD$\downarrow$)}
& \multirow{2}{*}{\shortstack{Lat.\\(ms)$\downarrow$}} \\
\cmidrule(lr){2-4}\cmidrule(lr){5-6}
& NYU & ScanNet & iBims & P3M-P & P3M-NP & \\
\midrule
\multicolumn{7}{l}{\emph{Is anything beyond the frozen body needed?}}\\
Head-only (body frozen)
& 44.30 & 43.55 & 48.16 & 180.97 & 170.92 & 47.7 \\
\midrule
\multicolumn{7}{l}{\emph{Prior or architecture?}}\\
Rand-init Full-FT
& 22.90 & 26.43 & 21.73 & 11.56 & 12.32 & 47.7 \\
\midrule
\multicolumn{7}{l}{\emph{Given the prior, does extra capacity or a generative interface help?}}\\
4-stage conv head (425K)
& 15.89 & 14.64 & 16.45 & 6.90 & 8.30 & 48.0 \\
Full fine-tuning (4B)
& 15.87 & \textbf{14.06} & 15.84 & 6.29 & 7.43 & 47.7 \\
Latent target
& 17.78 & 16.45 & 17.45 & 10.12 & 11.58 & 74.4 \\
VAE-frozen (pixel-sup.)
& 16.30 & 14.73 & 16.13 & 8.38 & 10.29 & 74.4 \\
Edit paradigm
& 15.94 & 14.37 & 15.83 & 6.68 & 8.00 & 118.1 \\
\rowcolor{oursdark}
\textbf{ReChannel-4B} (LoRA, Thin)
& \textbf{15.82} & 14.29 & \textbf{15.54} & \textbf{5.99} & \textbf{6.98} & \textbf{47.7} \\
\bottomrule
\end{tabular}
\end{table*}

\subsection{Diagnostic Ablations}
\label{sec:exp_ablation}

Table~\ref{tab:ablation_diagnostic} is organized as three diagnostic
questions, each isolated by perturbing one axis from our configuration
on a matched FLUX-Klein 4B backbone at $512^2$.

\textbf{LoRA adaptation is necessary.}
With the body fully frozen and only the linear head trained, normals
collapse to $44.30^\circ\,/\,43.55^\circ\,/\,48.16^\circ$ on
NYU\,/\,ScanNet\,/\,iBims, and matting plateaus at SAD
$180.97\,/\,170.92$ on P3M-P\,/\,NP, close to the trivial
constant-prediction regime. The pretrained RGB field is rich, but its
channels are not aligned with task-native quantities; lightweight LoRA
adaptation is needed before a linear readout becomes meaningful.

\textbf{A strong pretrained prior is necessary.}
Training the same architecture from random initialization leaves a
large gap on both tasks ($22.90^\circ$ on NYU normals, $+7.08^\circ$
over ours; SAD $11.56\,/\,12.32$ on matting, $\sim 2\times$ worse).
The readout thus relies on a strong pretrained prior, not on backbone
scale or token-grid structure alone. We do not claim generative
pretraining is uniquely required; our contribution concerns the output
interface, not the pretraining source.

\textbf{The adapted token field already carries the output; the interface only reads it.}
Once the DiT organizes the RGB observation into a task-aligned token
field, the target's spatial structure is already in place, so the
interface only has to read it out. Extra output-side machinery thus does
not help: a $13\times$ larger convolutional head and full fine-tuning of
the 4B body both fail to beat our token-local linear readout (full
fine-tuning only ties on one normal split, at far higher cost). Forcing
the target back through a generative interface---latent, VAE-decoded, or
edit---is consistently worse (Table~\ref{tab:ablation_diagnostic}),
re-rendering through a path the token field has made unnecessary. The
token-local readout is the most accurate configuration in the ablation.

\subsection{Efficiency}
\label{sec:exp_eff}

On a single L40S GPU at $512^2$, ReChannel runs at $47.7$ ms per image,
identical to a forward pass of the adapted T2I backbone: the token-local linear
head adds no measurable cost. The accuracy-matched alternatives all pay for a
target-side decode. The latent-target and VAE-frozen variants require a VAE
decoding pass and run at $74.4$ ms ($1.56\times$ slower), and the edit paradigm
runs at $118.1$ ms ($2.48\times$ slower) under the same backbone, resolution, and
LoRA rank. Removing the target-side VAE decode accounts for the $1.56\times$ gap
over the latent and VAE-frozen variants; the remaining gap up to $2.48\times$
over the edit paradigm additionally reflects its extra generative passes. Unlike
these baselines, ReChannel never runs the decoder at all: the target is read out
in the same forward pass that produces the token field, so accuracy and speed
come from the same design choice rather than trading off against each other. The
most accurate configuration in our study is thus also the fastest.

\section{Conclusion}
We revisited the output interface between text-to-image generative priors and
dense prediction. ReChannel removes the target-side VAE and treats the adapted
DiT token field as a spatial carrier from which task-native pixel patches are
read out by a token-local linear head. Across six tasks and more than a dozen
benchmarks, this simple interface reaches state-of-the-art results on
trimap-free matting, KITTI depth, and referring segmentation, while running up
to $2.48\times$ faster than an edit-plus-latent decode counterpart. These
results suggest that dense perception benefits from the RGB-native field
organized by generative pretraining, not from its target-side rendering
interface. Extending ReChannel beyond FLUX-Klein and pixel-aligned targets
remains future work.

\section*{Acknowledgements}
We thank Google's TPU Research Cloud (TRC) program for granting us access to Cloud TPUs.

\clearpage

\bibliographystyle{splncs04}
\bibliography{main}

\end{document}